\pdfoutput=1

\documentclass[11pt]{article}

\usepackage[preprint]{acl}

\usepackage{times}
\usepackage{latexsym}

\usepackage[T1]{fontenc}

\usepackage[utf8]{inputenc}

\usepackage{microtype}

\usepackage{inconsolata}

\usepackage{algorithm}
\usepackage{algorithmic} 
\definecolor{codegray}{gray}{0.9}
\newcommand{\code}[1]{%
  \begingroup\setlength{\fboxsep}{1pt}%
  \colorbox{codegray}{\texttt{\hspace*{2pt}\vphantom{Ay}#1\hspace*{2pt}}}%
  \endgroup
}
\usepackage{listings}
\definecolor{codegreen}{rgb}{0,0.6,0}
\definecolor{codegray}{rgb}{0.5,0.5,0.5}
\definecolor{codepurple}{rgb}{0.58,0,0.82}
\definecolor{backcolour}{rgb}{0.95,0.95,0.92}
\lstdefinestyle{mystyle}{
    backgroundcolor=\color{backcolour},   
    commentstyle=\color{codegreen},
    keywordstyle=\color{magenta},
    numberstyle=\tiny\color{codegray},
    stringstyle=\color{codepurple},
    basicstyle=\ttfamily\footnotesize,
    breakatwhitespace=false,         
    breaklines=true,                 
    captionpos=b,                    
    keepspaces=true,                 
    numbers=left,                    
    numbersep=5pt,                  
    showspaces=false,                
    showstringspaces=false,
    showtabs=false,                  
    tabsize=2
}
\usepackage{booktabs}
\usepackage{authblk}
\usepackage{graphicx}
\graphicspath{pics/}
\usepackage{caption}
\usepackage{boldline} 
\usepackage{multirow}
\usepackage{sidecap}
\usepackage{fancyvrb}
\usepackage{subcaption}
\usepackage{natbib}
\usepackage{amsmath}
\usepackage{amsfonts}
\usepackage{xcolor}
\usepackage{bm}
\usepackage{hyperref}
\usepackage{rotating}
\usepackage{adjustbox}
\usepackage{array}
\definecolor{green}{rgb}{0.0, 0.5, 0.0}
\definecolor{red}{rgb}{0.82, 0.1, 0.26}
\usepackage{pifont}

\newlength{\Oldarrayrulewidth}

\lstdefinelanguage{json}{
    basicstyle=\ttfamily,
    showstringspaces=false,
    breaklines=true,
    frame=lines,
    backgroundcolor=\color{lightgray},
    literate=
     *{0}{{{\color{blue}0}}}{1}
      {1}{{{\color{blue}1}}}{1}
      {2}{{{\color{blue}2}}}{1}
      {3}{{{\color{blue}3}}}{1}
      {4}{{{\color{blue}4}}}{1}
      {5}{{{\color{blue}5}}}{1}
      {6}{{{\color{blue}6}}}{1}
      {7}{{{\color{blue}7}}}{1}
      {8}{{{\color{blue}8}}}{1}
      {9}{{{\color{blue}9}}}{1}
      {:}{{{\color{red}{:}}}}{1}
      {,}{{{\color{red}{,}}}}{1}
      {\{}{{{\color{black}{\{}}}}{1}
      {\}}{{{\color{black}{\}}}}}{1}
      {[}{{{\color{black}{[}}}}{1}
      {]}{{{\color{black}{]}}}}{1},
}
\title{GRAMMAR: Grounded and Modular Methodology for Assessment of Closed-Domain Retrieval-Augmented Language Models}

\author{\textbf{Xinzhe Li}, \textbf{Ming Liu}, \textbf{Shang Gao}}
\affil{School of IT, Deakin University, Australia \\
\texttt{\{lixinzhe, m.liu,shang.gao\}@deakin.edu.au}}

\begin{document}
\maketitle

\begin{abstract}
Retrieval-Augmented Generation (RAG) systems are widely used across various industries for querying closed-domain and in-house knowledge bases. 
However, evaluating these systems presents significant challenges due to the private nature of closed-domain data and a scarcity of queries with verifiable ground truths.
Moreover, there is a lack of analytical methods to diagnose problematic modules and identify types of failure, such as those caused by knowledge deficits or issues with robustness.
To address these challenges, we introduce GRAMMAR (\textbf{GR}ounded \textbf{A}nd \textbf{M}odular \textbf{M}ethodology for \textbf{A}ssessment of \textbf{R}AG), an evaluation framework comprising a grounded data generation process
and an evaluation protocol that effectively pinpoints defective modules.
Our validation experiments reveal that 
GRAMMAR provides a reliable approach for identifying vulnerable modules and supports hypothesis testing for textual form vulnerabilities.
An open-source tool accompanying this framework is available in our GitHub repository \url{https://github.com/xinzhel/grammar}, allowing for easy reproduction of our results and enabling reliable and modular evaluation in closed-domain settings.
\end{abstract}

\begin{figure*}[ht!]
    \centering
    \includegraphics[width=0.95\textwidth]{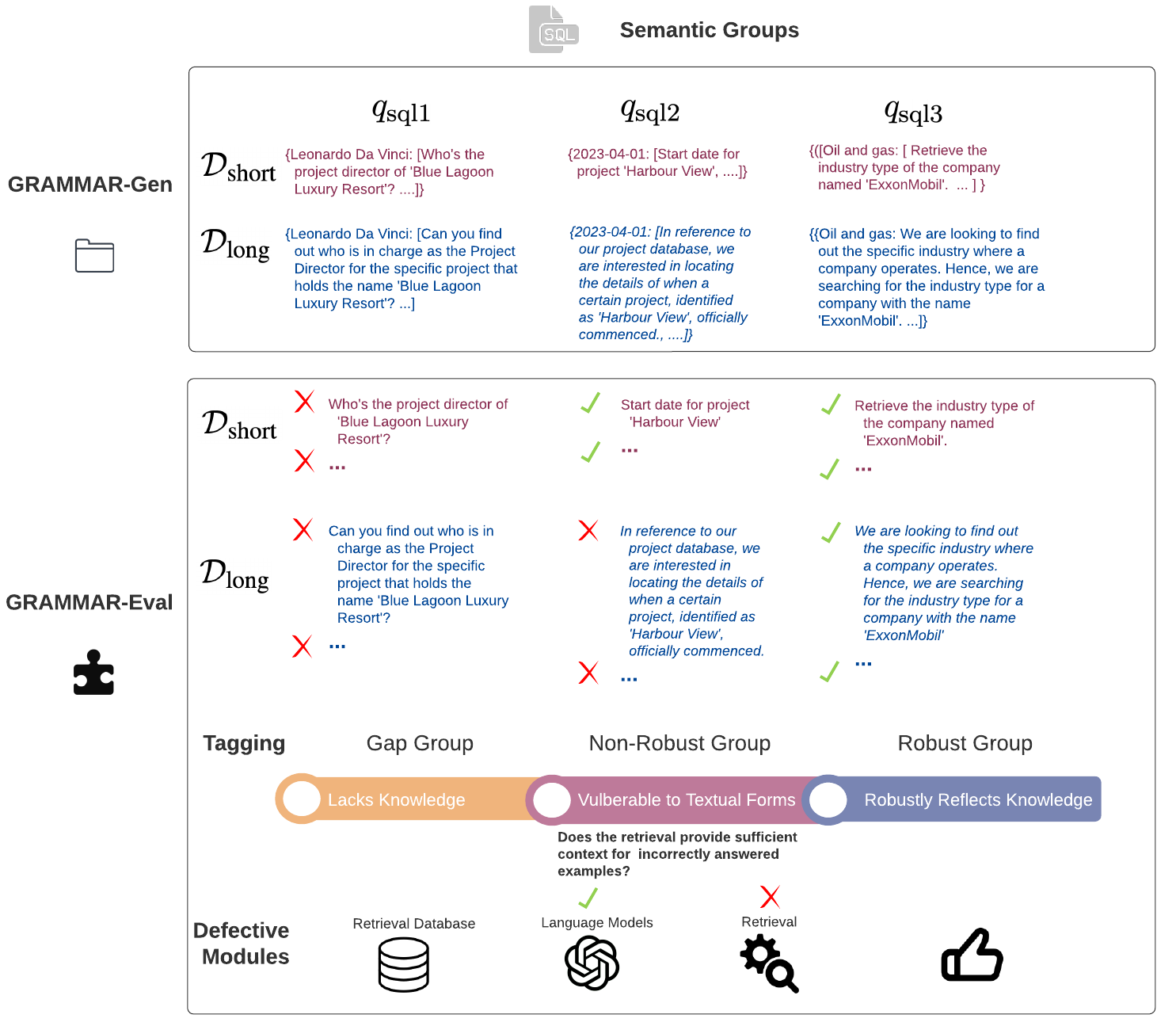}
    \caption{An Example of Applying the GRAMMAR Framework for Modular Evaluation and Hypothesis Testing. The upper section demonstrates the data generation process for creating sets of hypothetically robust ($\mathcal{D}{\text{robust}}$) and non-robust ($\mathcal{D}{\text{non-robust}}$) data. The lower section depicts the evaluation protocol that utilizes the generated data to identify defective modules and facilitate hypothesis testing.}
    \label{fig:grammar}
\end{figure*}

\section{Introduction}
The emergent capabilities of Large Language Models (LLMs) have driven significant research and the widespread deployment of Retrieval-Augmented Generation (RAG) systems for closed-domain settings. This growing trend highlights the critical need for reliable evaluation methods tailored to RAG systems.

A key challenge in evaluating RAG systems is the collection of domain-specific data that includes accurate ground truths. Researchers often rely on data originally designed for human assessment, as seen in studies such as \citet{santurkar2023whose, wang2022lsat, zhong-etal-2022-analytical, zhong2023agieval, hendrycks2021measuring, choi2023chatgpt}. However, these data sources typically reflect open-domain commonsense and world knowledge, whereas industrial in-house RAG systems require data that captures closed-domain knowledge, such as details about company projects and employees. Although user queries can be gathered during system deployment, obtaining accurate ground truths remains a significant challenge. To address this, recent studies have explored reference-free LLM evaluators for cases where ground truths are unavailable \citep{chern2023factool, min2023factscore, es2023ragas}. However, the reliability of these reference-free evaluation methods is still questionable.

Another critical challenge is assessing robustness in RAG systems, where a system may possess adequate knowledge for the semantics of a query but respond inconsistently to different query forms, as highlighted in \citet{shen2023chatgpt}. 
Developers and researchers often hypothesize about factors that could impact robustness, such as vulnerable modules (e.g., retrieval mechanisms or language models) or specific input attributes. However, existing research, e.g., studies on adversarial robustness \citep{alzantot2018generating, li-etal-2020-bert-attack, Li2019TextBuggerGA}, lacks effective tools for identifying non-robust modules and analytically testing these hypotheses.

Recognizing these challenges, the paper makes the following contributions:


\textbf{Proposing a grounded, controllable data generation process (GRAMMAR-Gen).}
To address the issue of ground truth scarcity, we propose GRAMMAR-Gen, a grounded data-generation process designed to ensure reliable evaluation. This process leverages relational databases and LLMs to extract ground truths via SQL queries. Additionally, the template-based query generation enables scalable data creation. By leveraging the capabilities of LLMs, it allows the controlled generation of data with diverse linguistic attributes.
Notably, a similar data-generation process for evaluating domain knowledge was proposed concurrently by \citet{tu2024r}\footnote{It was published after the initial preprint/submission of this work.}.

\textbf{Proposing a protocol for modular evaluation and hypothesis testing (GRAMMAR-Eval).}
GRAMMAR-Eval combines a grouping and tagging mechanism based on GRAMMAR-Gen with effective strategies to both identify knowledge gaps in the retrieval database and determine whether robustness issues stem from the retrieval mechanism or the language models (LMs). Our empirical results demonstrate the effectiveness of GRAMMAR-Eval in identifying deliberately bugged retrieval systems and accurately testing hypotheses about which input attributes contribute to robustness issues. An example of this process is illustrated in Figure~\ref{fig:grammar}.

\section{Background}
Evaluating the performance of Retrieval-Augmented Generation (RAG) systems, especially in closed-domain settings, poses unique challenges, leading to an absence of reliable reference-based evaluation protocols. This section outlines existing approaches and highlights the gaps addressed by our proposed framework, GRAMMAR.


\paragraph{Retrieval-augmented Generation (RAG)}
RAG systems augment LLMs with retrieval for domain-specific use where extensive databases can potentially provide necessary information.
A retrieval can be used for retrieving relevant passages or documents. A dense retrieval \citep{mialon2023augmented, lewis2020rag, borgeaud2022improving} consists of an embedding model to produce a vector \(v_q\) for any given query \(q\). This vector \(v_q\) is then used to identify pertinent text segments normally via dot product calculations with the vectors of text chunks. The most relevant chunks are aggregated to form an additional context \(c\), which serves as the factual basis for the response generation, until adding another chunk would surpass the maximum allowable context length. Subsequently, an LLM will process this aggregated context alongside the original query \(q\) to generate a predicted answer \(\hat{a}\) for it.

In contrast, sparse retrieval computes similarity without parametric embedding models. For instance, keyword-matching approaches \citep{manning2009introduction} select documents that share the highest number of common words for context retrieval. Additionally, methods like TF-IDF \citep{chen-etal-2017-reading} factor in the inverse document frequency, emphasizing the significance of less common words which are likely to be more indicative of pertinent content.


\paragraph{Reference-Free Evaluation}
Reference-free evaluation methods \citep{es2023ragas} assess model performance without requiring predefined ground truths. These methods can be categorized according to their applicable systems:
1) Protocols for LLMs: FactScore \citep{min2023factscore} and FacTool \citep{chern2023factool} use retrieval combined with LLMs as validators, but these are not directly applicable to RAG systems as retrieval is reserved for evaluation.
2) Protocols for RAG: ARES and RAGAS focus on evaluating the faithfulness of generated answers against retrieved contexts, with LLMs scoring the relevance and correctness of the context \citep{saadfalcon2023ares, es2023ragas}.
3) Protocols for Both: SelfCheck \citep{manakul2023selfcheckgpt} utilizes the stochastic nature of LLMs for self-validation.
While these methods offer some insights, their reliability is often questioned, especially for closed-domain applications where specific, accurate answers are crucial.

\paragraph{Reference-Based Evaluation}
While reference-free evaluation can be applied for evaluating both open-domain and closed-domain queries, reference-based evaluation typically assesses open-domain knowledge. The reason is that open-domain queries are widely available with ground-truth answers from public resources, exams, and surveys \citep{santurkar2023whose, wang2022lsat,zhong-etal-2022-analytical, zhong2023agieval, hendrycks2021measuring,choi2023chatgpt}. These evaluations compare generated answers to known correct answers, calculating metrics that reflect the model's accuracy.
However,  closed-domain scenarios lack reference-based evaluation due to the unavailability of ground-truth answers. GRAMMAR-Gen is proposed to solve this issue.

\section{Is Reference-free Evaluation Reliable?} \label{sec:reliability}
This section analyzes the reliability of reference-free evaluation methods using 198 examples from an engineering company. The RAG system for evaluation is implemented with dense retrieval and GPT-3.5 \footnote{An OpenAI embedding model for document embedding and \textit{gpt-3.5-turbo-16k}}. 

This preliminary study briefly highlights the potential benefits of reference-based evaluation, as our primary contribution focuses on proposing a reference-based evaluation framework. A detailed analysis of reference-free evaluation on state-of-the-art models is left for future work.

\paragraph{Two Evaluation Perspectives: Optimism and Cynicism }
To structure this analysis, reference-free evaluation is treated as a binary classification task. The ground truth reflects the actual correctness of RAG responses, while the evaluation models' judgments serve as the task's predictions. Thus, low precision indicates an optimistic bias (where incorrect RAG responses are often judged as correct), and low recall indicates a cynical bias (where correct RAG responses are often judged as incorrect).

\paragraph{Two Evaluation Protocols}
We assess two reference-free protocols. 
\begin{itemize}
    \item RAGAS-Fact \citep{es2023ragas}: This protocol utilizes the context-query-response triplets to assess the veracity of responses. It evaluates the faithfulness of a response by calculating the ratio of claims grounded on the context to the total claims made. This process involves identifying statements that hold atomic facts, following the methodology outlined by \citet{chern2023factool}.
    \item SelfCheck \citep{manakul2023selfcheckgpt}: 
    This protocol relies on the stochastic generation of responses, based on the premise that incorrect answers are unlikely to be consistently produced. This principle, initially applied to LLMs, is adapted in our analysis of RAG systems. In a departure from its original application, we enhance the evaluation prompt to include not only stochastic responses but also the query itself. Refer to Appendix \ref{app:llm_eval_prompts_full} for details. To generate four stochastic samples, we adjust the temperature setting to 1.0, contrasting with a temperature of 0.0 used to generate the primary response. A response is deemed correct if it aligns consistently across all stochastic samples.
\end{itemize}

\paragraph{Results: Both are Extremely Optimistic on Wrong Predictions While SelfCheck Becomes Too Cynical on Correct Predictions}
As shown in Table \ref{tab:reliability_reference_free_eval}, both RAGAS and SelfCheck achieve low precision (19\% and 15\%, respectively), highlighting a deficiency in correctly identifying erroneous predictions. For RAGAS, this may be because the RAG system produces responses that, while contextually relevant, fail to directly address the intended query. However, RAGAS demonstrates high recall, indicating accurate assessment of correct RAG responses.
Overall, the preliminary insight suggests that reference-free evaluation methods are not reliable for evaluation, underscoring the critical need for a robust, reference-based evaluation framework to ensure reliable model assessment.

\begin{table}[ht!]
    \centering
    \footnotesize
    \begin{tabular}{p{1.8cm}rr}
        \toprule
        \small
         & Precision & Recall \\
        \midrule
        RAGAS-Fact 
        & 19\% (11-27\%)   
        & 92\% (86-97\%) 
        \\\addlinespace 
        SelfCheck 
        & 15\% (0.08-0.22\%)
        & 50\% (0.4-0.59\%)
        \\ \bottomrule
    \end{tabular}
    \caption{Reliability of reference-free evaluation protocols.  Taking the size of the sample into account, the Z-test with 95\% confidence intervals is utilized. }
    \label{tab:reliability_reference_free_eval}
\end{table}

\section{GRAMMAR} \label{sec:grammar}
This section introduces GRAMMAR-Gen and GRAMMAR-Eval for grounded and modular evaluation:
1) GRAMMAR-Gen begins by generating templates based on a database schema, enabling scalable data creation \footnote{Further details can be found in Appendices \ref{app:possibilities_of_sql_templates} and \ref{app:scalable_breakdown}}. 
By leveraging LLMs, GRAMMAR-Gen offers controllability in generating text variations, which not only supports scalable data generation but also facilitates modular robustness evaluation and hypothesis testing. The process of SQL and text template generation is detailed in \S~\ref{sec:generating_templates}.
SQL queries, serving as intermediate representations, are used to create ground-truth answers by querying relational databases, as explained in  \S~\ref{sec:generate_eval_data}.  The overall process is illustrated in 
Algorithm~\ref{alg:query_generation} and Figure \ref{fig:scaled_generation}.
\textit{Note that the use of relational databases does not limit GRAMMAR's applicability in practice}, as specified in Appendix~\ref{app:generalizability}.
2) GRAMMAR-Eval, an evaluation protocol that effectively pinpoints defective modules by assessing the knowledge deficit of the retrieval database and detecting robustness issues originating from either the retrieval mechanisms or the language models. Further details can be found in \S~\ref{sec:modular_eval}.

\begin{algorithm}[tb]
	\caption{GRAMMAR-Gen} 
    \label{alg:query_generation}
    \small
    \textbf{Input}: SQL Template Generator $g_{\mathrm{sql}}$,
    Text Template Generator $g_{\mathrm{t}}$, Semantic criteria for SQL template generation $\mathcal{C}_{\mathrm{sql}}$, Linguistic criteria for text template generation $\mathcal{C}_{\mathrm{t}}$,
	Database $\mathcal{D}$,
    Database Schema $\mathcal{S}=\left\{S_1, S_2, \ldots, S_n\right\}$, where each $S_i$ is a schema for a table.\\
\textbf{Output}: Final evaluation data $Q$
\begin{algorithmic}[1] 
    \STATE $Q \gets \emptyset$ 
	\STATE  $\mathcal{S}_{\text{target}} \in \mathcal{S}$ 
    \STATE $\{tpl_{\text{sql}}\} \gets g_{\mathrm{sql}}(\mathcal{S}_{\text{target}}, \mathcal{C}_{\mathrm{sql}})$ 
        \FOR{$tpl_{\mathrm{sql}} \in \{tpl_{\text{sql}}\} $}
        \STATE $\{tpl_{\text{t}}\} \gets g_{\mathrm{t}}(tpl_{\mathrm{sql}}, \mathcal{C}_{\mathrm{t}})$ 

        \STATE $P \gets$ extract placeholders from $tpl_{\mathrm{sql}}$
        
        \FOR {each $p \in P$}
            \STATE $C \gets$ extract column from $p$
            \STATE $V_p \gets$ query $\mathcal{D}$ for distinct values of $C$
        \ENDFOR
        \STATE $Comb \gets$ Cartesian product of $\{V_p : p \in P\}$
        \FOR {each $comb \in Comb$}
            \STATE $q_{\mathrm{sql}}  \gets$ substitute $comb$ into $tpl_{\mathrm{sql}}$ 
            \STATE $a \gets$ query $\mathcal{D}(q_{\mathrm{sql}})$  
            \STATE  $\{q_t\}   \gets$ substitute $comb$ into $tpl_{\mathrm{t}} \in \{tpl_{\text{t}}\}$ 
            \STATE $Q \gets Q \cup \{ (\{q_t\}, a)\}$
        \ENDFOR
    \ENDFOR

    \STATE \textbf{return} $Q$
\end{algorithmic}
\end{algorithm}

\subsection{Generating Query Templates} 
\label{sec:generating_templates}
The GRAMMAR-Gen process first utilizes database schemas and LLMs to generate SQL query templates and their corresponding textual forms.

\paragraph{Database Schema}
Database schema serves as the blueprint that defines how classes of entities are organized and the relations among them. Appendix \ref{app:knowledge_repr_struc} demonstrates the reasons why we use database schema for knowledge representation.
Let a database schema with a collection of $N$ tables $S=\left\{S_1, S_2, \ldots, S_N \right\}$, where each $S_i$ is a schema for a database table. 
A table schema $S_i$ consists of the table name $T_i$, a set of attributes $\left\{A_1, A_2, \ldots, A_M\right\}$ and a set of constraints, i.e., $S_i=\left(T_i,\left\{A_1, A_2, \ldots, A_m\right\}\right., C )$, where the constraint $C$ demonstrates the primary key $PK$ and foreign keys $FK$. Specifically, $FK \left(T_i . A_k \rightarrow T_j . A_h\right)$ indicates a foreign key $A_k$ in table $T_i$ referencing the primary key $A_h$ in table $T_j$.
The schema defines the knowledge structure of an entity, which can be enough to generate query semantics that is not tied to any specific linguistic expressions and sensitive data.

\paragraph{Generating SQL Templates}
\label{sec:generating_sql_templates}
A SQL template $tpl_{\text{sql}}$ essentially represents the semantics of a query.  
Various query semantics are formulated through the application of relational algebra operators. In our study, we focus on three fundamental operators: Select $(\sigma)$, Project $(\pi)$, and Join $(\bowtie)$.
An automated SQL template generator $g_{sql}$ is powered by a generative LLM.
When given a target schema $\mathcal{S}{\text{target}}$ and specified SQL criteria $\mathcal{C}{\mathrm{sql}}$, $g_{\text{sql}}$ generates a set of SQL query templates.  

\paragraph{Generating Text Templates}
\label{sec:generate_text_templates}
\( t p l_{\mathrm{sql}} \) can be instantiated into various text templates, each denoted as $t p l_t$. This process, driven by the text template generator \( g_{\mathrm{t}} \), leverages linguistic criteria $\mathcal{C}_{\mathrm{t}}$ to transform \( t p l_{\mathrm{sql}} \) into natural language forms that align with specified linguistic characteristics, e.g., complexity, length and stylistic nuances \footnote{Examples of $\mathcal{C}_{\mathrm{sql}}$ and $\mathcal{C}_{\mathrm{t}}$ are specified in Appendix \ref{app:verbalizer}.}.  
The text template generator \( g_{\mathrm{t}} \) is also operationalized using a sophisticated language model, such as GPT-4, which is adept at producing diverse linguistic variations of the same semantic content. 
$$\{tpl_{\text{t}}\} = g_{\text{t}}(tpl_{\text{sql}}, \mathcal{C}_{\text{t}}),$$ where \( \{tpl_{\text{t}}\} \) signifies the resulting list of text templates.

\subsection{From Templates To Evaluation Data}
\label{sec:generate_eval_data}
Utilizing a database $\mathcal{D}$ that aligns with the predefined schema, the framework samples a diverse range of queries along with their corresponding ground-truth answers. This dataset, derived from SQL and text templates filled with database content, provides a rich source for evaluation.

\paragraph{Generating Queries Via Placeholder Fill-in}
The placeholder fill-in step populates the templates with actual data from the rows of tables in $\mathcal{D}$. Specifically, the approach employs a SELECT query format: ``SELECT DISTINCT \{column\_name\} FROM \{table\_name\};'' to ensure variety in the data points.
It handles SQL query templates with multiple placeholders by employing combinations of placeholders and using Cartesian products to generate multiple query permutations. This approach leads to a comprehensive set of SQL queries ($q_{\mathrm{sql}}$) and their natural language equivalents, the text queries ($q_{t}$).

\paragraph{Generating Answers} \label{sec:generate_answers}
Answer generation is integral to completing our dataset. Each SQL query ($q_{\mathrm{sql}}$) is executed against $\mathcal{D}$ to match with a factual answer $a$. Each answer is then paired with the relevant text queries $\{q_t\}$, forming the basis of our evaluation data.

\begin{figure*}[ht!]
    \centering
    \includegraphics[width=0.95\textwidth]{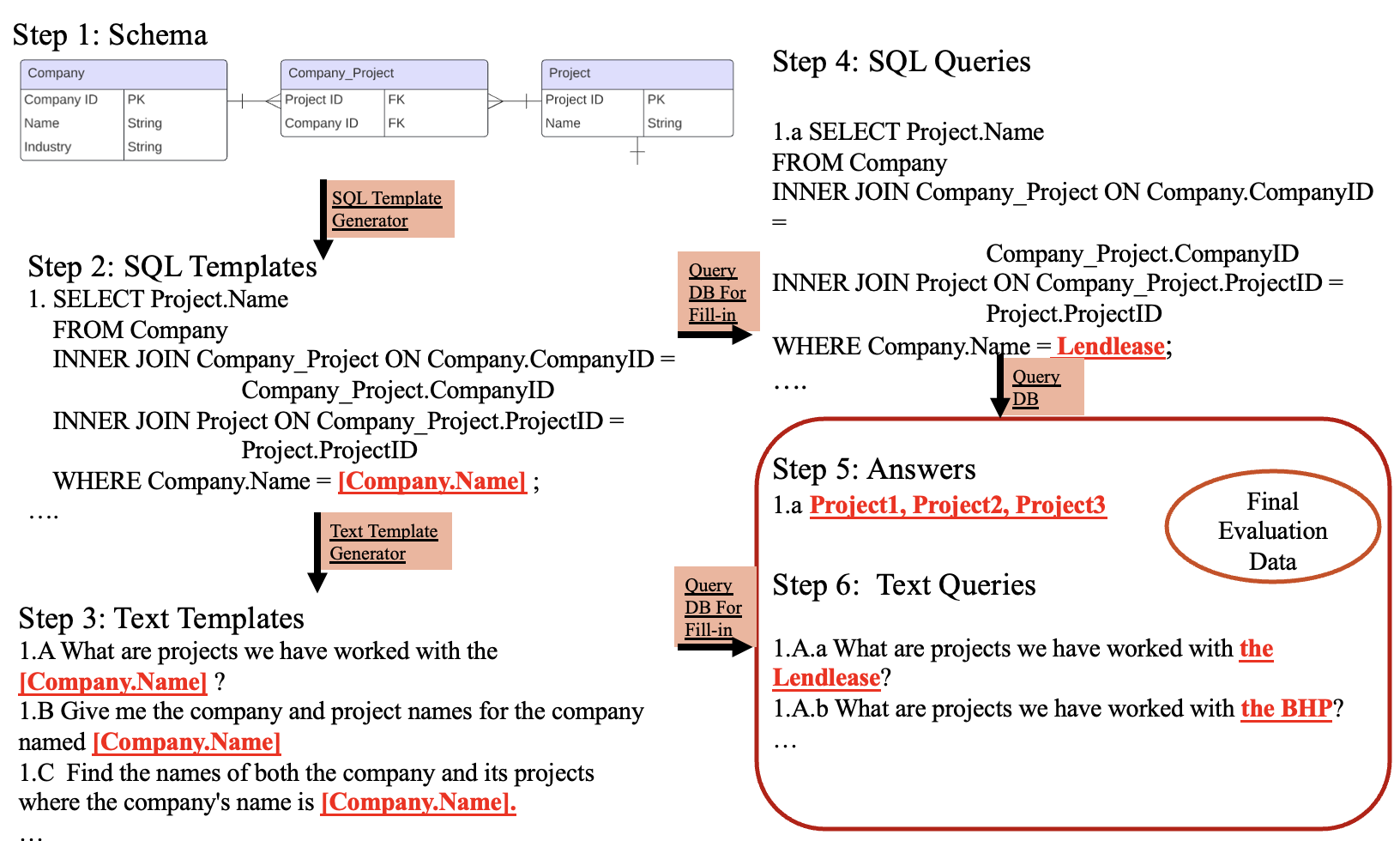}
    \caption{Scaled Generation of Query-answer Pairs. Step 3 and step 4 are independent of each other and depend only on the SQL templates.}
    \label{fig:scaled_generation}
\end{figure*}

\subsection{Modular Evaluation}
\label{sec:modular_eval}
Evaluating the overall performance of an RAG model can obscure the specific weaknesses or robustness of its constituent modules. This section introduces a modular evaluation protocol to address this. 

\paragraph{Query Grouping and Tagging}
The process involves grouping and tagging queries based on their semantics. Specifically, with GRAMMAR-Gen, textual queries $Q$ generated from a particular SQL query $q_{\mathrm{sql}}$ are organized into a semantic group.
Depending on the model $M$'s performance, $q_{\mathrm{sql}}$ (and consequently, the group $Q$) are tagged into three principal categories, as shown in Figure \ref{fig:grammar}:
\textbf{1) Gap Groups} emerge when the model $M$ consistently provides incorrect answers for every text query within a subset $S \subseteq Q$ produced through the data generation process. This reflects a deficiency in $M$ 's knowledge or capability. Formally, this scenario is described as:
$$
\forall q_t \in S, \neg M_e(M\left(q_t\right)),
$$
where $M_e$ is the evaluation model, which returns ``TRUE'' if the model's response $M\left(q_t\right)$ is correct.
\textbf{2) Robust Groups} are established when model \(M\) accurately responds to all text queries within the subset \(S\), evidencing a robust comprehension of the SQL logic. Formally, this is articulated as:
    $$ \forall q_t \in S, M_e(M(q_t)).$$
\textbf{3) Non-robust Groups} are characterized by model \(M\)'s ability to potentially answer the query semantics correctl, but with at least one query \(q_t\) within \(S\) that \(M\) fails to predict correctly. This is formally denoted as:
$$ \exists q_t \in S, M_e(M(q_t)) \land \exists q_t \in S, \neg M_e(M(q_t)).$$

\paragraph{Assessing Retrieval Database via Gap Groups} 
Gap examples, which are incorrectly predicted instances within gap groups, highlight errors in the retrieval database that arise due to its limited coverage. The adequacy of the knowledge within the database can be quantified as follows:

{
\small
\begin{equation} 
\text{Acc}_\text{retrieval\_db} = 1 - \frac{\text{Number of Gap Groups}}{\text{Total Number of Semantic Groups}}
\end{equation} 
}

A low $\text{Acc}_\text{retrieval\_db}$ suggests the need for expanding the database. 

\paragraph{Isolating Retrieval Database Errors}
To ensure accurate evaluation of the retrieval and language models without interference from retrieval database errors (i.e., distinguishing knowledge gaps from robustness performance), two strategies have been developed:
\textbf{1) Removing Gap Examples}: To assess the other two modules, we calculate refined accuracy, denoted as $R$, by excluding gap examples:
\begin{equation} 
\begin{split} 
R = \frac{\text{\# with Correct Predictions}}{\text{Total \#}- \text{Total \# in Gap Groups}},
\end{split}
\end{equation} 
where ``\#'' represents ``the number of instances''.
\textbf{2) Balancing Gap Groups}: To understand the advantage of this refined accuracy over baseline accuracy $\text{Acc}$, consider the following relationship: 
\begin{equation} 
\begin{split} 
\text{Acc} &= R \times \frac{\text{Total \#} - \text{Total \# in Gap Groups}}{\text{Total \#}}
\\
&= R \times (1 - \frac{ \text{Total \# in Gap Groups}}{\text{Total \#}})
\end{split}
\end{equation} 
This equation shows that a smaller normalized term, $\lambda = \frac{ \text{Total \# in Gap Groups}}{\text{Total \#}}$, enhances evaluation accuracy. Intuitively, a smaller $\lambda$ implies fewer gap examples (errors propagated from the initial module), leading to more accurate results. 
Maintaining a consistent $\lambda$ across evaluation models or datasets ensures accurate assessments, such as for the hypothetically robust dataset $\mathcal{D}_{\text{robust}}$ and the non-robust dataset $\mathcal{D}_{\text{non-robust}}$ discussed in \S \ref{sec:toy_case}. With GRAMMAR-Gen, an equal number of text queries can be generated within each semantic group by adjusting the text template generator.

\begin{table*}[!t]
    \centering
    \footnotesize
    \begin{tabular}{
    >{\raggedright}p{2cm}
    >{\raggedleft}p{1cm}
    >{\raggedleft}p{1cm}|
    >{\raggedleft}p{1cm} 
    >{\raggedleft}p{1cm}
    >{\raggedleft}p{1.5cm}
    >{\raggedleft}p{1.5cm}
    >{\raggedleft\arraybackslash}p{1.5cm}}
        \toprule
        & \multirow{2}{*}{\begin{tabular}{c}SQL\\Templates\end{tabular}} 
        & \multirow{2}{*}{\begin{tabular}{c}SQL\\Queries\end{tabular}} 
        & \multirow{2}{*}{\begin{tabular}{c}Text\\Templates\end{tabular}} 
        & \multicolumn{2}{c}{Text Queries} 
        & \multicolumn{2}{c}{Text Queries (Balanced)}   \\
        & & & 
        & Short 
        & Long
        & Short
        & Long
        \\ \midrule
        Aurp
       & 11
       & 157
       & 30 
       & 314 & 257 & 471 & 471
       \\ \addlinespace
       Spider-Open
       & 5
       & 57
       & 10
       & 397 & 436 & 570 & 570
       \\ \addlinespace
       Spider-Closed
       & 5
       & 57
       & 10 
       & 426 & 430 & 570  & 570
        \\ \bottomrule
    \end{tabular}
    \caption{Statistics of Evaluation Datasets}
    \label{tab:stat_data}
\end{table*}

\begin{table*}[ht!]
    \centering
    \footnotesize
    \begin{tabular}{llrrrrrr}
        \toprule
         \multicolumn{2}{c}{Isolating Errors from} 
         & \multicolumn{2}{c}{Aurp} 
         &  \multicolumn{2}{c}{Spider-Closed} 
         &  \multicolumn{2}{c}{Spider-Open} \\
         Retrieval Database & LLM 
         & Robust & Non-Robust 
         & Robust & Non-Robust 
         & Robust & Non-Robust  \\
         \midrule
         
         Baseline: No Action 
         & No Action
         & 0.27 & \textbf{0.51} 
         & 0.26 & \textbf{0.28} 
         & 0.67 & \textbf{0.87}
         \\ 
         & Context Comparison
         & 0.29 & \textbf{0.51} 
         & 0.31 & \textbf{0.34}
         & \textbf{0.37} & 0.35
         \\ \midrule
         
         Remove Gap Groups 
         & No Action
         & 0.95          & \textbf{0.96} 
         & \textbf{0.58} & 0.44 
         & \textbf{0.98} & 0.96
         \\ 
         & Context Comparison
         & \textbf{1} & 0.98 
         & \textbf{0.62} & 0.49
         & \textbf{0.51} & 0.39
         \\ \midrule
         
         Balance Gap Examples 
         & No Action
         & \textbf{0.27} & 0.27 
         & \textbf{0.28} & 0.21
         & \textbf{0.69} & 0.68
         \\ 
         & Context Comparison
         & \textbf{0.29} & 0.28 
         & \textbf{0.32} & 0.28 
         & \textbf{0.36} & 0.29
         \\ \bottomrule
    \end{tabular} 
    \caption{Validating GRAMMAR's tricks in in detecting non-robust retrieval module under a deliberately constructed RAG pipeline.  Accuracy is calculated on selected examples across various types of semantic groups.}
    \label{tab:modular_eval}
\end{table*}

\paragraph{Identifying Non-Robust Retrieval or Non-Robust LM: A ``Context Comparison'' Approach} 
To determine whether robustness issues stem from the retrieval module or the language model, we focus on incorrectly predicted queries within the non-robust group, each referred to as $q_\text{target}$.
Specifically, we pose the following question:

\emph{Does the retrieval module provide sufficient context for $q_\text{target}$?}

By leveraging other semantically identical but correctly answered queries within the group, we can assess whether the context provided is sufficient. Specifically, we compare the document indices $\text{idx}\text{target}$ associated with $q\text{target}$ to the indices of correctly answered queries, denoted as $I = \text{idx}1, \text{idx}2, \dots$. If $\text{idx}\text{target}$ matches any index in $I$, the context provided by $\text{idx}\text{target}$ is considered sufficient.

\section{A Toy Case for Validation: Detecting A Vulnerable Retrieval} \label{sec:toy_case}
This section demonstrates the effectiveness of GRAMMAR on question answering tasks.

\subsection{Experiment Setup}
\paragraph{Evaluation Target}
To validate our framework, we deliberately introduce a vulnerability in the retrieval component by using a basic keyword-matching retrieval approach. We use GPT3.5 for language generation. 
While this method is less common in practical applications, it is inherently weak in handling lengthy queries, making it an ideal candidate for testing our evaluation framework.

\paragraph{Evaluation Benchmarks and Datasets}
Closed-domain RAG systems lack established benchmarks, especially because questions must be answered using documents containing knowledge that is not publicly available or known to public LLMs. To address this gap, we modify the Spider benchmark \citep{yu-etal-2018-spider} and create two synthetic benchmarks with fictitious data.
\textbf{1) Context-Augmented Spider (Open-Domain)}: 
We use the ``company\_employee'' relational database from Spider to simulate a general real-world scenario. Documents for retrieval are synthesized using GPT-4 based on the database.
\textbf{2) Context-Augmented Spider (Closed-Domain)}: 
This is similar to the first dataset, but with company names modified to fictitious entities that are either unused by specific companies or generic according to the latest GPT-4 models.
\textbf{3) Aurp}: To create a strictly closed-domain evaluation, we synthesize fictitious facts about a fictional company called Aurp. A relational database and corresponding retrieval documents are generated so that questions derived from the database can only be answered by the provided documents.
Details on the data synthesis process are provided in Appendix~\ref{app:synthetic_data}, and the database schemas are outlined in Appendix~\ref{app:database_schema}. 
Using these three settings, datasets are generated by GRAMMAR-Gen to validate the design of GRAMMAR-Eval. The statistics of the generated datasets are summarized in Table~\ref{tab:stat_data}.
We have evaluated various perspectives on data quality to ensure the reliability of our framework, as detailed in Appendix~\ref{app:eval_data_quality}.

\subsection{Results}
This section validates the effectiveness of modular evaluation within the proposed framework in identifying defective modules.

\paragraph{GRAMMAR-Eval's Effectiveness on Modular Evaluation}
The results in Table~\ref{tab:modular_eval} highlight the value of modular evaluation in identifying the defective retrieval. 
Regardless of the methods used to isolate errors from the retrieval database and LLMs, the accuracy consistently identifies the vulnerable retrieval module. The next three paragraphs will provide a more detailed analysis.

\paragraph{Preventing Forward Error Propagation}
Preventing the propagation of knowledge deficits from retrieval databases is crucial. The two proposed strategies effectively address this issue. In all benchmarks, after either removing gap groups or balancing gap examples, accuracy on the robust text form consistently outperforms the non-robust text form (with the exception of Row 3 in Table \ref{tab:modular_eval}, where the results are very close).

\paragraph{Impact of Backward Errors from LLMs on Retrieval Assessment} 
In some cases, such as Row 3 in Table \ref{tab:modular_eval}, language model inaccuracies reduce the accuracy, masking the true performance of retrieval.  
While these instances are not prevalent in our dataset, they still distort the results.  
This distortion may be due to the stochastic nature of the LM's performance, or, in a less favorable scenario, because the non-robust text forms used in retrieval inadvertently benefit the LM. Based on the results from the other two datasets, it is more likely the former.

\paragraph{Applicability of GRAMMAR to Open-Domain RAG}
Table \ref{tab:modular_eval} also demonstrates the effectiveness of GRAMMAR for performing modular evaluation on the open-domain Spider dataset. 
However, in open-domain settings, the GPT model may generate accurate answers from its internal knowledge. This can lead to inaccuracies in assessing both the retrieval system's knowledge gaps and the robustness of the retrieval mechanism.
Specifically, the potential limitations in applying GRAMMAR-Eval to open-domain settings include:
\textbf{1) Misjudgment of the Retrieval Database}: Some gap examples in the retrieval database may go undetected, leading to an inflated $\text{Acc}\text{retrieval\_db}$. This issue is empirically demonstrated in Table~\ref{tab:open_domain_effect}, which compares $\text{Acc}\text{retrieval\_db}$ before and after filtering out open-domain queries for Spider-Open. 
Closed-domain queries are identified by applying GRAMMAR to detect gap groups using answers from the LLM-only system.
\textbf{2) Limitations of Context Comparison}: Context comparison may fail when correctly answered examples do not actually receive sufficient context. 
This issue is evident in Table \ref{tab:modular_eval}, where the accuracy decreased under ``Context Comparison'' compared to ``No Action.'' The decrease in accuracy is likely due to the introduction of incorrect context, which adds noise to the language model's generation, thereby hindering its performance.
In contrast, accuracy consistently improves for the two closed-domain benchmarks when context comparison is applied.
\begin{table}[!ht]
    \centering
    \footnotesize
    \begin{tabular}{lr}
    \toprule
    & $\text{Acc}_\text{retrieval\_db}$ 
    \\ \midrule
    With Open-Domain Queries     & 0.53 
    \\ \addlinespace
    Without Open-Domain Queries  & 0 
    \\\bottomrule
    \end{tabular}
    \caption{Wrong judgement on retrieval database with open-domain queries}
    \label{tab:open_domain_effect}
\end{table}

\section{Conclusions}
In this work, we have introduced the GRAMMAR framework specifically designed for Retrieval-Augmented Generation (RAG) systems in closed-domain settings. 
The data generation module (GRAMMAR-Gen) improves the reliability of evaluation, while the evaluation protocol (GRAMMAR-Eval) enables detailed modular analysis.
 
\paragraph{Reproducibility}
Our open-source Python package enables easy reproduction of all results, including the regeneration of experimental data and outcomes from our validation study. Detailed instructions on using the package are provided in Appendix \ref{app:reproduce}. 
Due to data privacy policies, the 198 examples and the RAG system from the sponsoring company are not included.


\section{Limitations}
The data generation process has certain limitations, such as its inability to handle scenarios with multiple correct answers (refer to Appendix \ref{app:multiple_correct_answers}). Additionally, the expressiveness of the generated queries is constrained by the database schema and SQL (refer to Appendix \ref{app:expressiveness}), limiting the ability to generate queries that require multi-step reasoning or free-form responses. However, as demonstrated in the case study, simple queries are sufficient for identifying and addressing intrinsic robustness issues within the models.

Besides, while the simple context comparison method performs well for closed-domain evaluation, it may need refinement for more complex retrieval tasks. This method was primarily used to validate the importance of modular evaluation and to establish a foundation, along with a reusable code base, for future development.


\bibliography{anthology,acl_latex}
\appendix
\section{Prompt Templates for LLM-Based Evaluation } \label{app:llm_eval_prompts_full}
For SelfCheck, given that our responses are typically concise, we opt for assessing the overall response rather than rather than performing a sentence-by-sentence validation as in the original implementation, as shown in Table \ref{tab:llm_eval_prompts_full}.
\begin{table*}[ht!]
    \centering
    \footnotesize
    \begin{tabular}{p{14cm}}
    \toprule
    \textbf{\textit{SelfCheck}} from \citep{manakul2023selfcheckgpt} \\
    Context: \{context\} \newline
    Sentence: \{sentence\} \newline
    Is the sentence supported by the context above? \newline
    Answer Yes or No: 
    \textbf{\textit{SelfCheck-QA}} \\
    Query: \{query\}  \newline
    Answer A: \{answer\}  \newline
    Answer B: \{stochastic\_answer\} \newline
    Do both answers address the query with equivalent meaning? \newline
    Use only ``Yes'' or ``No'' for your evaluation: 
  
    \\ \hline
    \textbf{\textit{Ragas}} from \citep{es2023ragas} \\
    Natural language inference. Use only 'Yes' (1), 'No' (0) and 'Null' (-1) as verdict. \newline 
    context: \{context\} \newline 
    statement: \{statement\} \newline 
    verdict:  
    
    \\ \hline
    \textbf{\textit{Ours}} \\
    Evaluate the accuracy of the given response in relation to the true answer for the specified query. After evaluating, provide a judgement as either ``Correct'' or ``Incorrect'' based on whether the \#\#Given Response\#\# accurately matches the \#\#True Answer\#\#.\newline
    \#\#Query\#\#: \{query\} \newline
    \#\#True Answer\#\#: \{true\_answer\} \newline
    \#\#Given Response\#\#: \{given\_response\} \newline
    \#\#Judgement\#\#:
    \\ \bottomrule
    
    \end{tabular}
    \caption{Prompt templates for LLM-based Evaluation.}
    \label{tab:llm_eval_prompts_full}
\end{table*}
\section{Knowledge Structure and Representation} \label{app:knowledge_repr_struc}
Commonly, the knowledge is formalized by entities along with their relations, e.g., a knowledge graph, ontology and a database schema. Instead of using knowledge graphs containing concrete entities and their relations, we use the schema-based definition, where the schema is characterized by entity types (classes of entities). 
An entity type is defined by its name, its attributes and relations with other entity types that will be concretized as components in a database, i.e., the table name, columns and foreign keys. 

The definition of entity types can provide combinatorial expansion of data generation, protecting the leak of private information (normally stored as rows of tables) during the data generation process and the ability of retrieving ground-truths via Structured Query Language (SQL). The three advantages are the reasons why we represent knowledge in database schema rather than knowledge graphs. It disentangles the private data stored in the database with universal metadata expressed in the database schema. Since schemas and templates contain only meta-data for structuring the real data, there is no private issue to use commercial large language models for this process.

\section{Generator Criteria}
\label{app:verbalizer}
Each criterion for the SQL generator and text generator ($\mathcal{C}_{\mathrm{sql}}$ and $\mathcal{C}_{\mathrm{t}}$, respectively) consists of brief instructions provided to the LLM, specifying the type of SQL or text that needs to be generated.

Table \ref{tab:criteria_example} provides examples:
\begin{table*}[ht!]
    \centering
    \footnotesize
    \begin{tabular}{p{14cm}}
    \toprule
    \textit{$\mathcal{C}_{\mathrm{sql}}$: Generating SQL queries with one placeholder }  \\
    \begin{Verbatim}[commandchars=\\\{\}]
    Each query must contain at least one parameter placeholder in the WHERE clause. 
    \end{Verbatim}

    \textit{$\mathcal{C}_{\mathrm{t}}$: Generating short queries}  \\
    \begin{Verbatim}[commandchars=\\\{\}]
    Short and Clear: Keep your queries short and straightforward. 
    Cut down on words and skip parts of speech, such as conjunctions and articles.
    It's okay to use fragmented phrases as long as they still convey the full meaning. 
    Valid examples: "client of '[Project.Name]'" or "client for '[Project.Name]'"; 
    Invalid Examples: "Find the client of a project named '[Project.Name]'. 
    \end{Verbatim}
                
    \textit{$\mathcal{C}_{\mathrm{t}}$: Generating long queries} \\
    \begin{Verbatim}[commandchars=\\\{\}]
    Complex Sentence Structure: Ensure your queries are always in complete sentences. 
    Opt for longer, more complex sentence structures, 
    incorporating elements of speech like conjunctions and articles for fuller expression. 
    Each query should be at least 30 words long. 
    You can add context and background information to the query.
    \end{Verbatim}
    
    \\ \bottomrule
    
    \end{tabular}
    \caption{Examples of criteria for the SQL generator and text generator.}
    \label{tab:criteria_example}
\end{table*}

\begin{table*}[!ht]
    \centering
    \footnotesize
    \begin{tabular}{p{15cm}}
     \toprule
    \begin{Verbatim}[commandchars=\\\{\}]
    You are a SQL query Template Generator: Generate ACCEPTABLE SQL query 
    templates with placeholders according to the give data schema and 
    requirements. A simple example of an acceptable SQL query template is: 
    SELECT Industry FROM Company WHERE Name = '[Company.Name]';
    
    You must follow the basic criteria below except for other requirements: 
    ##CRITERIA##
    - The placeholder format should be a combination of a table name and 
    a column name, enclosed within square brackets, e.g., '[User.Name]'.
    - Use only 'SELECT' queries.
    - Select specific column(s) instead of using '*'. Avoid projecting 
    attributes that appear in the predicate.
    - The selected and condition columns in the query MUST BE MEANINGFUL and
    DESCRIPTIVE to ensure the queries are easily understood by non-technical 
    users.
    - Avoid using technical column names that don't clearly signify the nature 
    of the entities or objects involved, e.g., column for semantically void 
    record identifiers.
    - Do not create redundant or semantically duplicated queries when translated 
    into natural language.
    - Each query must contain at least one parameter placeholder in the WHERE 
    clause.
    - Ensure the query yields a specific and singular answer to avoid multiplicity 
    issues, thus facilitating accurate chatbot evaluation.
     \{SPECIFIC_REQUIREMENTS\}
    - If no acceptable SQL template can be generated with the given table and 
    column information, do not generate any text.

    ##RESPONSE FORMAT##
    - Output each SQL template as a single line, without any prefix or suffix.
    - Do not include any other text in your response, even something like
    ##RESPONSE_END##.

    ##DATA SCHEMA##
    \{GIVEN_SCHEMA\}

    ##RESPONSE_START##
    \end{Verbatim}
    
    \\ \bottomrule
     \end{tabular}
    \caption{A prompt template designed to guide the LLM in functioning as a controllable SQL template generator}
    \label{tab:sql_template_generator}
\end{table*}

\section{Details of SQL Template Generation}
\label{app:sql_template_generator}

\subsection{Constraints/Guidelines for Creating SQL Templates}
\label{app:constraints_guidelines}
The constraints limit the types of templates that can be generated. This section demonstrates two important constraints, which are verbalized as the prompt for LLMs. 

\paragraph{Selecting Attributes That Are Understandable to Humans}
The SQL template generator is required to follow the constraint: ``The selected and condition columns in the query MUST BE MEANINGFUL and DESCRIPTIVE to ensure the queries are easily understood by non-technical users.''.

\paragraph{Avoiding Answer Multiplicity}
\label{app:multiple_correct_answers}
In the evaluation of question-answering (QA) models, a unique challenge arises from the existence of multiple valid answers to a single query, which necessitates a nuanced approach to assessing model performance. Consider the question: ``Get the name of the client associated with the project named Innovation Precinct'' For such a question, a set of correct responses could include any combination of names from a predefined list, such as \{Apple, Amazon, Meta, Facebook\}. This multiplicity of correct answers underscores the complexity of evaluating QA models, as it requires the assessment mechanism to recognize and validate the full spectrum of possible correct answers rather than comparing the model's output against a single 'gold standard' answer. This scenario demands a more flexible method for answer validation that can accommodate the variability in correct responses. Also, evaluation metrics may be required to effectively measure the performance of QA models in handling diverse and equally valid answers. Such metrics must account for the exhaustive set of correct answers and evaluate the model's ability to retrieve any or all valid responses within the context of the query, thereby ensuring a comprehensive assessment of the model's understanding and response generation capabilities.
Generally, encompassing all valid responses within the database is not feasible. Hence, rather than let the non-robust results affect the accuracy of evaluation, we look for an approach to avoid it, i.e.,
only generating queries ensuring one ground-truth answer. Specifically, the evaluation of certain SQL queries, such as ``SELECT Name, Department, BusinessAddress FROM Employee WHERE JobTitle = 'designer''', requires a complete and thorough listing of multiple answers that can be dynamically changed. To solve the issue, the solution involves adding a specific criterion to the prompt for generating SQL queries. This criterion, ``Ensure the query yields a specific and singular answer'', aims to produce queries that result in a single, clear answer, thereby avoiding the complexities of multiple possible correct answers. 
For example, with this criterion, the query ``SELECT Industry FROM Company WHERE Name = '[Company.Name]';'' is preferred as it's likely to yield a singular answer about a company's industry based on a specific company name. In contrast, without this criterion, queries like ``SELECT Name FROM Company WHERE Industry = '[Company.Industry]';'' are acceptable but may result in multiple names, leading to evaluation difficulties due to data completeness and query multiplicity issues.

\subsection{Possibilities Of SQL Templates} \label{app:possibilities_of_sql_templates}
\paragraph{One Table/Entity}
Given a schema with only one table, the possibilities for SQL templates can be analyzed by considering the basic SQL operators like SELECT, WHERE. Here's a breakdown: 
\begin{itemize}
    \item Variation in Selected Attributes (SELECT):
    The number of SQL templates varies based on the combination of columns selected.
    If the ``Company'' table has $n$ columns, then theoretically, there are $2^n - 1$ possible combinations of columns for selection (excluding the case where no column is selected).
    \item Conditions in Queries (WHERE):
    Each SQL query can include zero or more conditions in the WHERE clause.
    The number of possible conditions is determined by the number of columns, the type of each column (text, numeric, date, etc.), and the range of operators applicable to these types (like =, <, >, LIKE, IN for text columns; =, !=, <, >, BETWEEN for numeric columns). The complexity increases combinatorially with multiple conditions combined using AND/OR.
\end{itemize}

\paragraph{Two Entities}
The relations between tables further amplifies the number of possible templates. 
For example, with two entities, ``Company'' and ``Project,'' and their associative table ``Company\_Project'', the possibilities for SQL templates expand significantly due to the introduction of joins and more complex WHERE clauses. Let's break down the possibilities. 
\begin{itemize}
    \item Selection Variations (SELECT):
    The number of SQL templates grows with the combination of columns selected across the three tables: ``Company'', ``Project'' and ``Company\_Project''.
    If ``Company'' has n columns, ``Project'' has m columns, and ``Company\_Project'' has p columns, the possible combinations for selection are $(2^n - 1) \times (2^m - 1) \times (2^p - 1)$.
    \item Join Conditions (JOIN):
    The introduction of the associative table ``Company\_Project'' allows for meaningful JOIN operations between ``Company'' and ``Project.''
    Templates can include joins like Company JOIN Company\_Project ON condition and Project JOIN Company\_Project ON condition, or a multi-table join linking all three.
    The variety of JOIN conditions adds another layer of complexity to the possible templates.
    \item WHERE Clause Complexity:
    With more tables, the WHERE clause can include a wider range of conditions, potentially involving attributes from any of the three tables.
    The complexity increases with the number of columns and their types across all tables, and combinations of these conditions.
\end{itemize}
    
    


\subsection{Limitations}
\paragraph{Scope of Using Automated SQL Generators}
The effectiveness of queries generated by LLMs hinges on the meaningfulness of table and column names to accurately reflect the essence of the entities or concepts they represent. If these names lack contextual clarity, the resulting queries may be impractical. To mitigate this, practitioners can conduct an informal assessment or consult established benchmarks that evaluate LLMs' proficiency in domain knowledge and commonsense reasoning, for instance, \citep{zhong2023agieval,hendrycks2021measuring}.

\paragraph{Queries Beyond SQL Expressiveness} \label{app:expressiveness}
While SQL and relational algebra offer a wide range of operations enabling the formulation of numerous user queries, certain semantic nuances exceed the expressive capabilities of SQL. The following examples illustrate such limitations:
\begin{itemize}
    \item Real-world queries often contain ambiguity and subjective interpretations that SQL struggles to accommodate. For instance, the term ``major'' in the query ``What are some major rail projects we've been involved with?'' does not directly translate into SQL criteria without additional interpretative steps.
    \item SQL queries are typically structured to elicit specific, predefined responses, contrasting with the open-ended nature of many real-world inquiries that seek exploratory or comprehensive answers.
\end{itemize}

\section{Breakdown of Scalable Data Generation} \label{app:scalable_breakdown}
The scalability of the proposed data generation process is achieved through two key factors: 1) the creation of templates using various SQL formulations from a modest, manually crafted database with a limited number of attributes and rows, and 2) the generation of text queries with LLMs. The potential number of query permutations can be expressed succinctly by the formula (see Figure \ref{fig:scaled_generation} for a visual representation):
\begin{equation}
\text{Total Query Variations} = M \times N \times Q,
\end{equation}
where ``M'' denotes the array of SQL templates that can be extracted from a given schema, ``N'' refers to the vast number of text templates that can be produced from a single SQL template, showcasing the flexibility of natural language, and ``Q'' accounts for the broad spectrum of text queries that can be generated from a single text template, with this diversity arising from different combinations of unique values filling the placeholders.

Let's break down each step of the proposed data generation process:
\begin{itemize}
    \item 1 schema $\Rightarrow$ M SQL Templates:
    The number of possible SQL templates (M) that can be generated from a defined schema is influenced by the SQL operators used, the number of tables, and the columns in the database schema.
    Each combination of tables and columns, along with different SQL operators (like SELECT, WHERE, JOIN), can lead to a unique SQL query template, e.g., the exponential complexity of the SELECT operator, and the combinatorial increase of possible predicates (See Appendix \ref{app:possibilities_of_sql_templates} for details). 
    Note that while the theoretical maximum is high, practical and meaningful queries will be a smaller subset, as specified in the case study and Appendix \ref{app:sql_template_generator}.
    \item 1 SQL Template $\Rightarrow$ N Text Templates:
    For each SQL template, there can be an arbitrary number (N) of textual expressions.
    This variation arises from the different ways to linguistically express the same SQL query due to the flexibility and richness of natural language, e.g., the variation of synonyms and sentence structures.
    \item 1 Text Template $\Rightarrow$ Q Text Queries:
    Each text template can lead to a number of text queries (Q), depending on the unique values available for each placeholder.
    With more than one placeholder, the potential for growth in the number of text queries is combinatorial, barring semantic conflicts (context conflict), where certain combinations of column values may not be semantically valid.
    Overall, this framework demonstrates a combinatorial expansion at each transition stage, especially notable in steps involving natural language due to its inherent variability. However, this growth is tempered by practical constraints such as the meaningfulness of queries (semantic validity) and the actual data distribution in the database. The exponential increase is most pronounced in the transition from text templates to text queries, where the permutations of placeholders can lead to a vast array of unique query possibilities.
\end{itemize}

\section{Use of Relational Databases Does Not Limit GRAMMAR's Applicability in Practice}
\label{app:generalizability}

The key point is that SQL databases can be easily self-constructed with a limited understanding of the RAG system's use cases to construct a simple schema (e.g., Figure 3) and insert a small number of rows into the tables (e.g., only 49 rows are manually inserted across 3 Aurp tables). Our approach is highly scalable for generating evaluation data, as shown in Table 2 and Figure 2, and highlighted in Line 241 with further details in Appendix E.

The requirements of defining a basic schema or inserting rows can be easily handled either by developers who normally have concrete ideas during RAG system development, by probing data during short-term testing of the RAG system before deployment, or even through a brief (1-hour) interview with users.

\section{Database Schema Overview} \label{app:database_schema}
\begin{figure*}[ht!]
    \centering
    \footnotesize
    \begin{subfigure}[b]{0.49\textwidth}
        \centering
        \includegraphics[width=\textwidth]{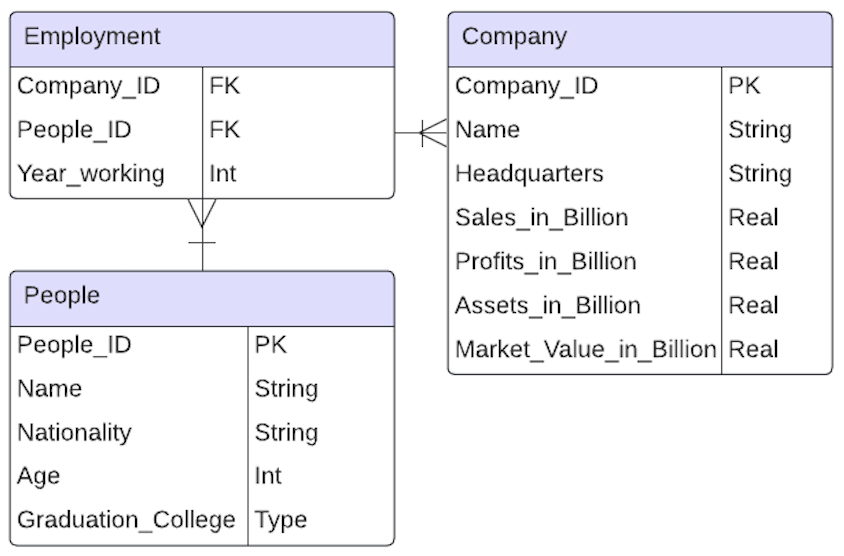}
        \caption{Spider.}
        \label{fig:db_schema_spider}
    \end{subfigure}
    \hfill 
    \begin{subfigure}[b]{0.49\textwidth}
        \centering
        \includegraphics[width=\textwidth]{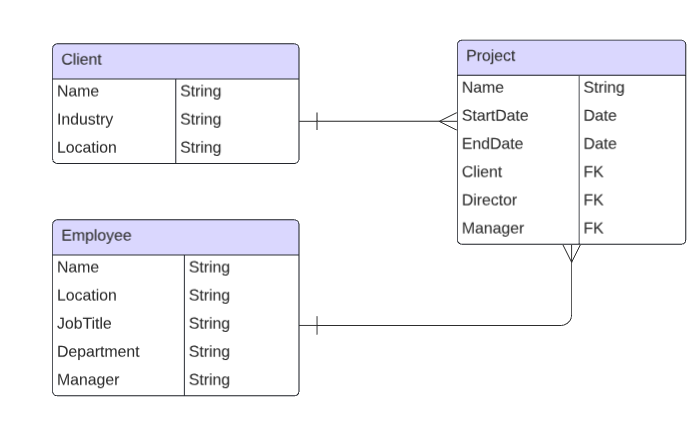}
        \caption{Aurp.}
        \label{fig:db_schema_aurp}
    \end{subfigure}
    \caption{Entity-Relationship Diagrams for Data Generation}
    \label{fig:db_schemas}
\end{figure*}

Figures~\ref{fig:db_schema_spider} and ~\ref{fig:db_schema_aurp} illustrate the database schemas used for an actual industrial context and a fabricated scenario, respectively, within our research.

\section{Synthetic Processes for Data Generation} \label{app:synthetic_data}

\subsection{Aurp Setup}
\label{app:aurp}
The generation starts from a company profile, i.e., a fictitious company named Aurp, and then goes to structural knowledge to create a relational database, including organizational structures, employees, clients and projects. Finally, all the information above is used to generate synthetic project documents. Below is a detailed process.
\begin{enumerate}
    \item \textbf{Generating Company Profile}: Initially, a company profile for Aurp is created, detailing its foundation year, headquarters location, CEO, number of employees, and the services it offers, such as bespoke architectural solutions, sustainable urban planning, and structural health monitoring.
    
    \item \textbf{Generating Organizational Structure}: Next, a project-oriented organizational structure is established, naming key positions and employees within the company, similar to real-world firms. This includes a wide range of roles from executive positions to specialized engineers and support staff.
    
    \item \textbf{Generating Employee Information}: For each employee listed in the organizational structure, detailed job titles, departments, and direct supervisors or managers are fabricated, creating a network of relationships and reporting lines within the company.
    
    \item \textbf{Generating Client Information}: The process creates a list of ten clients across various industries—ranging from technology and real estate to hospitality and healthcare—each with specified locations, thereby illustrating the company's diverse portfolio.
    
    \item \textbf{Generating Projects}: Specific projects are then devised, including names, locations, start and end dates, clients, project directors, and project managers. This step integrates previously generated data (client and employee information) to create realistic project scenarios.
    
    \item \textbf{Generating Project Reports}: This step involves synthesizing data from the previous steps to produce detailed analyses, updates, and outcomes of the various projects, which may or may not contain detailed information for clients and employees.
\end{enumerate}

\subsection{Context-Augmented Spider}
Since relational databases have been constructed in Spider, simple company and employee profiles with factual knowledge are manually crafted using the stringified templates in Table \ref{tab:spider_profiles}. The simple profiles are then used to prompt GPT-4 to generate Wikipedia-style pages containing 200-400 words.
Knowledge required for queries is sparse in all documents.

\begin{table}[!ht]
    \begin{minipage}{\columnwidth}
    \raggedright
    \begin{tabular}{p{3.5cm}}
        \begin{scriptsize}
        \begin{Verbatim}[commandchars=\\\{\}]
Company Profile
**Name**: \{company\_name\}
**Headquarters**: \{headquarter\}
**Industry**: \{industry\}
**Sales in Billion**: \{sales\}
**Profits in Billion**: \{profits\}
**Assets in Billion**: \{assets\}
**Market Value in Billion**: \{market\_value\}

Employee Profile
**Name**: \{name\}
**Age**: \{age\}
**Nationality**: \{nationality\}
**Graduation College**: \{graduation\_college\}
        \end{Verbatim}
        \end{scriptsize}
    \end{tabular}
    \end{minipage}
    \caption{Templates for company and employee profiles. Placeholders in curly brackets will be replaced by factual information in Spider databases.}
    \label{tab:spider_profiles}
\end{table}

\section{Evaluation of Data Quality}
\label{app:eval_data_quality}

\paragraph{Evaluation Perspectives}
To make our evaluation legitimate in terms of data quality, the following conditions should be satisfied:
\begin{itemize}
    \item Perspective 1: The generated text queries ask about the correct fact
    \item Perspective 2: The generated documents (for retrieval) contain those facts
    \item Perspective 3: Coherence and grammatical correctness are important aspects of data quality. However, it is widely acknowledged that GPT-4 models consistently generate coherent text \citep{coyne2023analyzing}, and our random manual checks further confirm this.
\end{itemize}

\paragraph{No Human Preferences in Factual QA Data}
Our setting is focused on factual QA tasks, where answers are either correct or incorrect, leaving no room for subjective human preferences.

\subsection{Evaluating Perspective 1, 2}
To verify these two points, a Machine Reading Comprehension (MRC) task is setup, where the generated document is given as context to answer the questions encoding the facts it generate on. 
In this setup, the queries can only be answered correctly if
1)the given MRC context (i.e., generated documents) contains correct facts, and
2) the queries do ask about the facts (since the MRC answers tend to be incorrect given the large amount of information in the documents).

These two points correspond to verifying condition 2 and 1, respectively. In simple terms, if the MRC model (or humans) consistently generates correct answers, the data quality satisfies conditions 1 and 2.

\paragraph{Empirical Results}
Using the balanced, short-query version of Spider (closed-domain) and Aurp datasets, where correct documents are given as context, the MRC model based on GPT-3.5 generates correct answers 80.11\% and 98.52\% of the time, respectively. This only confirms the data quality to some extent. 

\paragraph{Further Analysis}
To ensure a full reliability, we did error checking on the remaining 19.89\% and 1.48\% examples and check whether they undermine GRAMMAR’s reliability. The results show that:
All incorrect predictions resulted from failing condition 2: the lack of expected facts in the retrieval documents.
However, this does not undermine GRAMMAR’s reliability, since these examples were all correctly identified as gap examples.

\section{Reproducing Data and Results}
\label{app:reproduce}

\subsection{Reproducing Data}
The following steps illustrate how to use our open-source tool to reproduce the data. The datasets are also available along with the tool.
\begin{lstlisting}[language=Python, numbers=none]
from grammar.db_tool import DBTool
from grammar.llm import AnyOpenAILLM
from grammar.sql_template_generator import SQLTemplateGenerator
from grammar.text_template_generator import TextTemplateGenerator
from grammar.qa_generator import QADataGenerator


llm = AnyOpenAILLM(model_name = "gpt4-short") 

setup_env = "aurp"
if setup_env == "spider" or setup_env == "spider_closed":
    database_name = 'spider'
    connection_string = f'sqlite:///{database_name}/rel_database/company_employee.sqlite'
    schemas = [('company',), ('people',), ('company', 'people')]
elif setup_env == "aurp":
    database_name = 'Aurp'
    connection_string = "mysql+pymysql://root:password@localhost:3306/Aurp"
    schemas = [('client',), ('employee',), ('project', )]
db_tool = DBTool(connection_string)
\end{lstlisting}

\begin{lstlisting}[language=Python,numbers=none]
# Step 1: Generate SQL Query Templates
file_path = f"{setup_env}/SQLTemplateGenerator/sql_templates.json"
sql_template_generator = SQLTemplateGenerator.from_file(file_path, sql_connection=connection_string, llm=llm)
entities_to_sql_templates = sql_template_generator.generate_batch(schemas, override=False, verbose=True)
sql_templates = [tpl for entity, tpls in entities_to_sql_templates.items() for tpl in tpls]
\end{lstlisting}
Below are examples of generated SQL templates in JSON format, where the templates are keyed by table names.
\begin{lstlisting}[language=json,numbers=none]
{
    "('client',)": [
        "SELECT Location FROM Client WHERE Name = '[Client.Name]';",
        "SELECT Industry FROM Client WHERE Name = '[Client.Name]';"
    ],
    "('employee',)": [
        "SELECT JobTitle FROM Employee WHERE Name = '[Employee.Name]';",
        "SELECT Department FROM Employee WHERE Name = '[Employee.Name]';",
        "SELECT SupervisorOrManager FROM Employee WHERE Name = '[Employee.Name]';"
    ],
    ...
}
\end{lstlisting}

\begin{lstlisting}[language=Python, numbers=none]
# Step 2: Generate Text Query Templates
linguistic_attr = "long"
file_path = f'{setup_env}/TextTemplateGenerator/{linguistic_attr}.json'
text_template_generator = TextTemplateGenerator.from_file(file_path=file_path, verbalize_attrs=linguistic_attr, llm=llm) # Load existing generations to avoid re-generation
sql_to_text_templates = text_template_generator.generate_batch(sql_templates, verbose=True, num_generations=3, override=False)
text_template_generator.save(file_path=file_path, override=True)
\end{lstlisting}
Below are examples of generated text templates saved in JSON format.
\begin{lstlisting}[language=json, numbers=none]
{
    "SELECT StartDate FROM Project WHERE Name = '[Project.Name]';": [
        "Start date for project '[Project.Name]'",
        "Look up start date of '[Project.Name]'",
        "Get '[Project.Name]' start date"
    ],
    ...
}
\end{lstlisting}

\begin{lstlisting}[language=Python, numbers=none]
# Step 3: Generate Evaluation Data (Text Queries and Answers)
save_file = f"{linguistic_attr}.json"
qa_generator = QADataGenerator(db_tool)
all_answers_to_text_queries = qa_generator.generate(sql_to_text_templates)
qa_generator.save(all_answers_to_text_queries, database_name, save_file, overwrite=True)
\end{lstlisting}
Below are examples of generated query-answer pairs saved in JSON format.
\begin{lstlisting}[language=json, numbers=none]
[
    [
        "[('Maldives',)]",
        [
            "Get 'Blue Horizon Hotels' location details",
            "Find which location 'Blue Horizon Hotels' is located in",
            "Determine the location of 'Blue Horizon Hotels'"
        ]
    ],
    ...
]
\end{lstlisting}

\subsection{Reproducing Experiment Results}
The tool aslo provides two high-level Python objects to reproduce results from our experiments: \code{TaggedGroup} to perform tagging on each semantic groups and generate metrics for modular evaluation. The code below demonstrates an example to reproduce the results in Table \ref{tab:modular_eval}. The input to \code{get\_eval\_results} and \code{TaggedGroup} is a list of \code{RAGResult} objects, which is a data class including a query, the ground-truth answer and the RAG response.

\begin{lstlisting}[language=Python, numbers=none]
import json
from grammar.eval.result import RAGResult
from grammar.eval.tag_group import TaggedGroup 
from grammar.eval.match import SemanticsMatch

def get_eval_results(eval_results, linguistic_attr, root_dir, file_path):
    tagged_group = TaggedGroup(eval_results)
    semnatics_match = SemanticsMatch.from_file(root_dir=root_dir, verbalize_attrs=linguistic_attr)

    for eval_result in eval_results:
        # sleep for 20 seconds after 9 examples
        # if results.index(result) % 9 == 0 and results.index(result) != 0:
        #     print("Sleeping for 20 seconds")
        #     time.sleep(20)
        #     print("Waking up")
        eval_result.judge_retrieval_response(tagged_group=tagged_group, method='use_exist')
        eval_result.judge_rag_response(semnatics_match)

    num_retrieval_failure = sum([result.retrieval_judgement==0 for result in eval_results])
    print(f"Retrieval failed in {num_retrieval_failure} out of {len(eval_results)} examples")
    num_rag_failure = sum([result.judgement=="Incorrect" for result in eval_results])
    print(f"RAG failed in {num_rag_failure} out of {len(eval_results)} examples")
    semnatics_match.save(root_dir=f'{root_dir}', override=True)
    # semnatics_match.llm.gpt_usage_record.write_usage(model_name='chatgptk' )

    # save results
    results = [result.asdict() for result in eval_results]
    # ensure json serializable
    for result in results:
        result['true_document_ids'] = list(result['true_document_ids'])
        result['retrieved_document_ids'] = list(result['retrieved_document_ids'])
    with open(file_path, 'w') as f:
        json.dump(results, f, indent=4)

    return eval_results, tagged_group

root_dir = 'aurp'
closed_domain = True
results, metric = get_eval_results( 'short', root_dir, file_path=f'{root_dir}/eval_results/results_short_balanced.json')

# re-produce metrics in Table 3
print('Baseline Accuracy: ', metric.get_accuracy())
print('Accuracy (Remove LLM Errors): ', metric.get_accuracy(for_retrieval=True))
print('Removing Gap Examples: ', metric.get_robustness())
print('Removing LLM Errors & Gap Examples: ', metric.get_robustness(for_retrieval=True))
\end{lstlisting}

\end{document}